\DocumentMetadata{}
\documentclass[sigconf,review]{acmart}

\makeatletter
\def\mdseries@tt{m}
\makeatother

\pdfoutput=1

\usepackage{amsmath}
\usepackage{amsfonts}
\usepackage{graphicx}
\usepackage{booktabs}
\usepackage{array}
\usepackage{url}
\usepackage{listings}
\usepackage{xcolor}
\usepackage{caption}
\usepackage{subcaption}
\usepackage{float}
\usepackage{geometry}
\usepackage{algorithm}
\usepackage{algorithmic}
\usepackage{multirow}
\usepackage{mdframed}
\usepackage{xstring}
\usepackage{pifont}

\theoremstyle{definition}
\newtheorem{definition}{Definition}[section]

\usepackage{tikz}
\definecolor{GoodGreen}{RGB}{0,139,69}
\definecolor{BadRed}{RGB}{204,0,0}

\newcommand{\impact}[2]{%
  \IfStrEq{#2}{red}{%
    \textcolor{BadRed}{\textbf{#1}}%
  }{%
    \IfStrEq{#2}{green}{%
      \textcolor{GoodGreen}{\textbf{#1}}%
    }{%
      \textbf{#1}%
    }%
  }%
}

\begin{document}
\title[Terminus-4B]{Terminus-4B: Can a Smaller Model Replace Frontier LLMs at Agentic Execution Tasks?}

\author{Spandan Garg}
\email{spgarg@microsoft.com}
\affiliation{%
  \institution{Microsoft}
  \country{USA}
}
\author{Vikram Nitin}
\email{vikramnitin@microsoft.com}
\affiliation{%
  \institution{Microsoft}
  \country{USA}
}
\author{Yufan Huang}
\email{yufanhuang@microsoft.com}
\affiliation{%
  \institution{Microsoft}
  \country{USA}
}

\date{}

\begin{abstract}
Modern coding agents increasingly delegate specialized subtasks to subagents, which are smaller, focused agentic loops that handle narrow responsibilities like search, debugging or terminal execution. This architectural pattern keeps the main agent's context window clean by isolating verbose outputs (e.g. build logs, test results, etc.) within the subagent context. Typically when agents employ subagents for such tasks, they use frontier models as these subagents. In this paper, we investigate whether a finetuned small language model (SLM) can achieve comparable performance to frontier models in the task of agentic terminal execution. We present Terminus-4B, which is a post-trained Qwen3-4B model via Supervised Finetuning (SFT) and Reinforcement Learning (RL) using rubric-based LLM-as-judge reward, specifically for this task. In our extensive evaluation spanning various frontier models, training ablations and main agent configurations, we find that Terminus-4B is able to \textbf{reduce the main agent's token usage by up to $\sim$30\%} compared to the No Subagent baseline with no impact to agent performance on benchmarks like SWE-Bench Pro, TerminalBench 2.0 and our internal SWE-Bench C\# benchmark, which tends to be heavy in verbose execution tasks. Furthermore, Terminus-4B improves key metrics showing the main agent relying on the outputs of the subagent and doing fewer terminal execution tasks by itself. We see that our model not only closes the gap between the Vanilla Qwen model and frontier models like Claude Sonnet / Opus / GPT-5.3-Codex, but often even exceeds their performance.

\end{abstract}

\maketitle

\thispagestyle{empty}
\pagestyle{plain}

\section{Introduction}
In recent times, coding agents~\cite{vscode_2025, wang2024opendevinopenplatformai, claude_code_2024, sweagent} have made strides in software engineering tasks, ranging all the way from writing tests to resolving complex repo-level GitHub issues. A crucial component to all these workflows is terminal execution i.e. the process of running builds, installing dependencies and running diagnostic tests to reproduce issues and validate fixes. While essential, these tasks flood the agent's context window with terminal outputs. 
%Often these commands are run near the start of the trajectory and the outputs become a permanent fixture in the trajectory, costing more tokens and money with each subsequent request. 
A single verbose test can easily produce tens of thousands of tokens displacing the code context, editing and planning the agent needs to do for downstream decisions. 
As the trajectory grows, this creates a compounding problem: each new command output further crowds the context window, limiting how many actual problem solving steps the agent can take before it hits the token window limit. In practice, terminal output is often the single largest consumer of context in coding agent trajectories. We argue that this direct execution pattern, where the agent runs commands in its own context and absorbs the full output, is primitive and represents a major inefficiency in agent design of today's coding agents. The information the main agent needs from these terminal execution is often just a concise summary of what took place such as a line representing the error or a table at the end, and not the full raw output from the shell command. %Surprisingly, coding agents are designed to include this full context in the trajectory, which is highly wasteful. 

Popular coding agents have increasingly adopted a subagent architecture as a way to address context limitation of LLMs. Rather than performing expensive tasks in the main agent loop, the agent delegates them to a specialized subagent. The subagent executes the task independently based on the inputs provided and returns output in a format expected by the main agent, absorbing the context hit from the intermediate steps. This pattern has been successfully adapted to tasks like search~\cite{claude_code_2024}, debugging~\cite{debug2fix}, etc. We believe terminal execution is a natural candidate for this pattern.
%, since the outputs are among the most verbose artifacts in coding workflow, yet the information needed by the main agent is a short chunk of the full output.
Despite the clear fit, applying the subagent pattern to terminal execution has its own challenges. The subagent must be able to handle diverse repositories spanning various languages. It must be able to interpret errors correctly and decide what to do in response, handle timeouts gracefully, etc. Even more importantly, it should produce effective summaries as its final response, as that is the only window the main agent has into what the subagent did. A subagent producing vague summaries (e.g. \textit{"The build failed."}) or hallucinated outputs is worse than no subagent at all. This is because the main agent would not only need to repeat the work, but could be misled by a faulty subagent. Typically, subagents rely on frontier models because of their adaptability to different tasks. Given the narrow nature of this task, we believe that using an expensive frontier LLM is overkill and a much smaller model should be capable of performing the task. 

In this work, we apply the subagent pattern to terminal execution, by adding an \textit{Execution Subagent} to an existing agent. The subagent is given a single \textit{Terminal} tool, bounded by a turn limit and a targeted system prompt instructing the agent to return structured summaries. The main agent can delegate tasks to this subagent with a simple query. To avoid using an expensive frontier LLM, we introduce Terminus-4B, which is a Qwen3-4B~\cite{qwen} model post-trained specifically for agentic terminal execution. We conduct a two-stage training pipeline with supervised finetuning (SFT) on trajectories collected from our internal usage telemetry data, followed by Reinforcement Learning (RL) with Group Relative Policy Optimization (GRPO)~\cite{grpo} with a novel subagent training pipeline and rubric-based LLM-as-judge reward~\cite{llmjudge, rubric}. The subagent pipeline is able to isolate the subagent task from the main agent to ensure each rollout begins with the same problem, while keeping the main agent's role in the rollout minimal, resulting in cost-effective training. The reward scores candidate rollouts against reference trajectories across various quality dimensions and failure modes, providing a rich multi-dimensional training signal suited for this task. Our contributions are as follows:

\begin{itemize}

    \item \textbf{Execution Subagent \& Terminus-4B.} We design and integrate Execution Subagent into a production coding agent framework. By using our own custom finetuned Terminus-4B model with Execution Subagent, we can reduce token usage by up to $\sim$30\% while preserving the agent performance on challenging coding benchmarks like SWE-Bench Pro, TerminalBench 2.0 and our internal, SWE-Bench C\#.
    \item \textbf{Subagent Training Framework} We present a novel post-training framework for subagent model training that decouples the subagent from the main agent loop. This enables fast and cost-effective rollouts with minimal reliance on frontier LLMs. We are able to use a vanilla Qwen3-4B itself as the main agent during rollouts.
    \item \textbf{Task Formulation \& Reward Design} We introduce a novel rubric-based multi-dimensional LLM-as-judge reward that compares the rollouts against reference trajectories along suality and failure dimensions, providing a rich training signal for a task where traditional outcome-based rewards~\cite{instructgpt} aren't readily available.
    \item \textbf{Comprehensive Evaluation} We conduct an extensive ablation on the subagent design and model training using SWE-Bench Pro, TerminalBench 2.0 and our internal benchmark, SWE-Bench C\#, measuring impact on resolve rate, token efficiency and behavior signals. We complement this analysis with a 5-dimension LLM judge evaluation of the subagent response from the main agent's perspective. In our evaluation, we demonstrate that Terminus-4B matches or exceeds frontier model performance as an Execution Subagent.
\end{itemize}

\section{Background and Related Work} 
Our work intersects and extends several research threads in coding agent space, such as multi-agent / subagent architectures, small language models for agentic workloads, RL for multi-turn LLM-based agents and LLM-judge / rubric-based reward design. We discuss each below.

\subsection{Multi-Agent and Subagent Architectures}
Decomposing complex tasks across multiple agents has been studied extensively. AutoGen~\cite{autogen} provides a flexible framework for agent to agent conversation. Works like MetaGPT~\cite{metagpt}, ChatDev~\cite{chatdev}, explore role-based collaboration between multiple agents. He et al.~\cite{hesurvey} systematically reviewed the landscape of LLM-based multi-agent systems for software engineering highlighting the current capabilities and limitations of these approaches. Anthropic's multi-agent research~\cite{anthropicmultiagent} system adopts the orchestrator-worker pattern, where a lead agent delegates tasks to specialized subagents that operate in isolated contexts. Claude Code~\cite{claude_code_2024} further formalizes this pattern as subagents, with built in general purpose and plan subagents and the ability to author custom ones. Our Execution Subagent follows the same orchestrator-worker pattern, but specializes in the domain of terminal execution tasks within coding agents, a task where verbose tool outputs are especially prevalent.

\subsection{Small Language Models for Agentic Tasks}
A growing body of work argues that small language models are essential for the future of Agentic AI~\cite{slmfuture}. They argue that a bulk of agent invocations involve repetitive tasks for which SLMs are not only sufficient, but also 10-30 times cheaper than frontier LLMs. The Qwen3 family~\cite{qwen} represents a strong open-weight SLM family that has native tool-calling capabilities and growing body of research showing that appropriate post-training~\cite{deepseekr1, dangreason} can allow them to achieve competitive results on focused tasks. Terminal-4B applies these learnings and principles to the targeted yet impactful task of terminal execution.

\subsection{Terminal Tasks and Execution Agents}
Probably the closest to our domain, TerminalBench~\cite{terminal} provides a benchmark of realistic command-line tasks executed in sandboxed Docker environments, finding that Frontier LLMs still resolve fewer than 65\% of tasks while smaller models score only around 15\%. Recent coding-agent training works~\cite{qwencodernext, skyrlagent} explicitly evaluate on Terminal-Bench as an out of domain task to see if the LLM generalizes to such tasks. More closely related to our work, Gandhi et al.~\cite{endlessterminals} procedurally generate terminal tasks and use vanilla PPO to train small models for terminal use. Our work differs from this because we train via a novel rubric-based reward design on tasks mined from GitHub issues and treat terminal execution as a delegated subagent as our goal is to reduce main agent's token usage. To the best of our knowledge, prior work has treated terminal execution as a capability of the main agent rather than something that can be delegated to specialized subagent running a finetuned smaller model. Our work fills this gap.

\subsection{Context Management for Long-Horizon Tasks}
As LLMs and agents grow in capabilities, so do their trajectories in length and context management becomes an important concern when it comes to agent design. Long context inflates cost and degrades reasoning capabilities by diluting attention away from task-relevant tokens in the trajectory. Focus~\cite{focus} introduces an agent that autonomously decides when to consolidate key learnings into a persistent blocks and actively prunes the agent interaction history. Sun et al. introduce Context folding~\cite{contextfolding}, which is a framework that allows agents to branch sub-trajectories to handle subtasks and then fold them into the main trajectory. Memex(RL)~\cite{memex} introduces a compact representation of context with concise structured summaries and stable indices, SWE-ContextBench~\cite{swecontextbench} explicitly evaluates how summarized and raw context affects coding agents capabilities. Our subagent approach is complementary to these past approaches. Rather than compressing context, the Execution Subagent has an independent agentic loop to run verbose terminal commands and prevents their output from ever entering the main agent context, absorbing it in the subagent context and returning only a structured summary.

\section{Motivating Example}
\begin{figure*}
    \centering
    \includegraphics[width=0.95\linewidth]{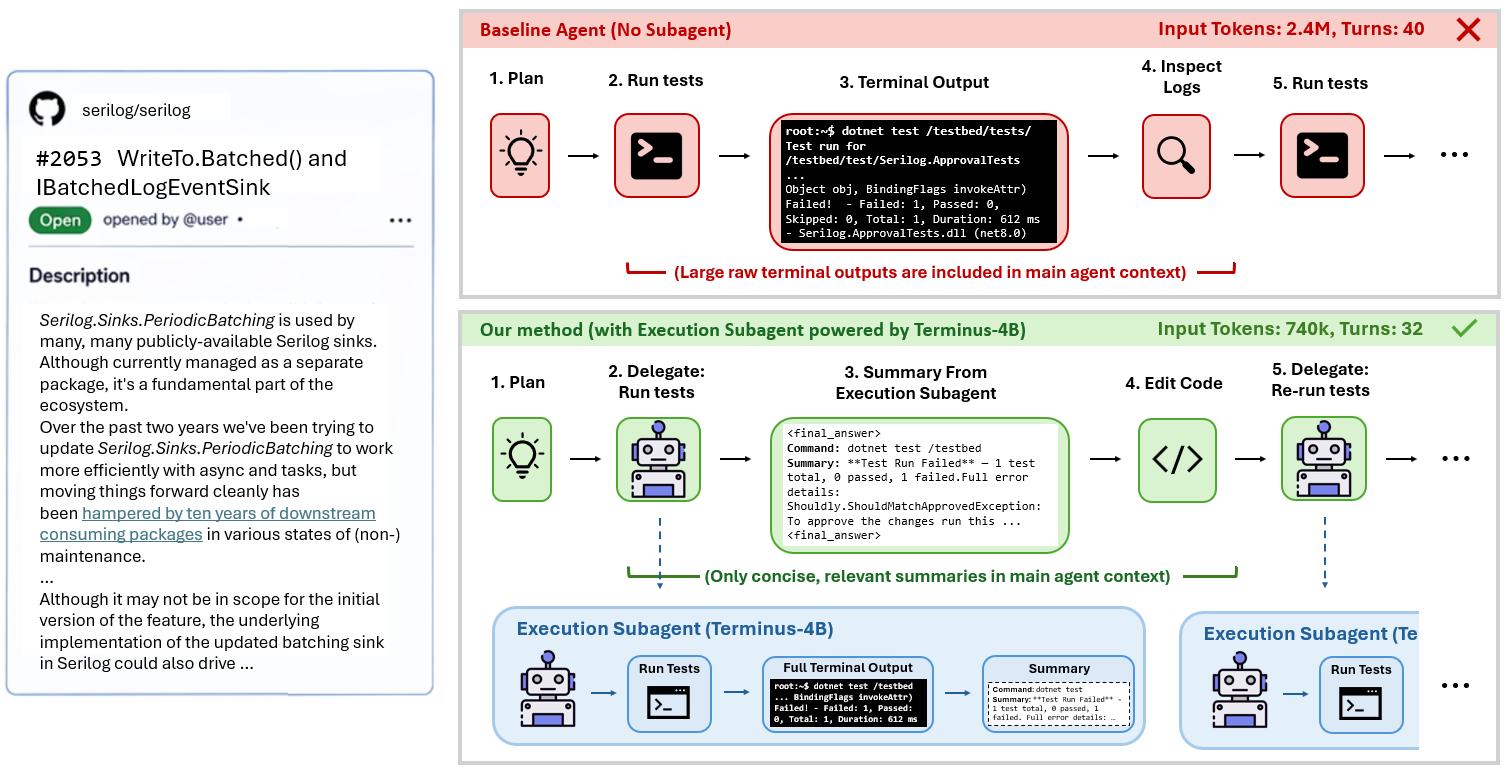}
    \caption{Contrasting agent trajectories for a real issue in the \texttt{Serilog} repository, with and without the Execution Subagent. Without the subagent (top), the main agent must directly invoke terminal commands, process verbose build and test output, and spend additional turns interpreting results, consuming significantly more tokens and turns. With our subagent (bottom), terminal
   execution tasks are delegated to the subagent, which absorbs the raw output and returns only the key findings in a concise final answer with a predefined format. The main agent never has to process the verbose terminal output, thereby preserving its context window.}
    \label{motivating_example}
\end{figure*}

In this section, we illustrate the benefits of our approach through a real world example. Figure~\ref{motivating_example} shows two agent trajectories solving the same task from the Serilog C\# repo (\texttt{serilog/serilog \#2053}). The issue requires the agent to add a new batching API surface. Since this a brand new feature, the agent must build the solution, run unit tests, identify failures and apply necessary fixes to make the build / test errors go away.

When we compare the trajectories output by the agent with and without the subagent, we see very different outcomes. In the baseline agent, each terminal command returns its full raw output into the agent's context window itself. Over the course of the trajectory, the main agent makes 18 direct terminal calls, many of which are repeated with slight variations as the agent tries different combinations of \texttt{grep} and \texttt{tail} filters on the output to extract the relevant information from the verbose build and test logs. Example:

\begin{lstlisting}[basicstyle=\ttfamily\tiny]
$ dotnet test ... 2>&1 | tail -40
$ dotnet test ... 2>&1 | grep -E "passed|failed|error"
$ dotnet test ... 2>&1 | grep -A 10 "Error Message"
$ dotnet test ... 2>&1 | grep -E "passed|failed|Error Message|Assert"
$ dotnet test ... 2>&1 | tail -20
...
\end{lstlisting}

Each call adds new raw output to the context, which gets sent as input context to the LLM for every subsequent turn. By the end of trajectory, the agent has consumed \textbf{2.46M} main agent tokens across 40 turns, with terminal output accounting for the majority of context.

In the subagent configuration, the main agent delegates the same task to the Execution Subagent (using Terminus-4B, our RL-trained model) with a single query: \texttt{``Run the build, then the unit and approval tests, and report pass/fail counts with error details''}. The subagent executes a total of 9 commands internally in the subagent context and returns a final answers that look like this: 
\begin{lstlisting}[basicstyle=\ttfamily\tiny]
<final_answer>
Command: dotnet build /testbed/Serilog.sln
Summary: Build succeeded. 9 warnings, 0 errors.

Command: dotnet test .../Serilog.Tests.csproj
Summary: All 769 tests passed, 0 failed, 0 skipped.

Command: dotnet test .../Serilog.ApprovalTests.csproj
Summary: Test Run Failed - 1 failed.
  Error: Serilog.received.txt does not match Serilog.approved.txt.
  The diff shows new API surface in LoggerAuditSinkConfiguration.
  To fix: update the approved snapshot file.
</final_answer>
\end{lstlisting}

Instead of the full raw terminal output, the main agent only receives a $\sim$200-token summary at a time, which immediately identifies the underlying issue causing the test failure. Overall the trajectory costs only 740k main agent tokens across 32 turns, which is significantly less compared to the baseline agent. Furthermore, since the subagent uses a Terminus-4B, the tokens consumed cost a fraction of the cost of main agent tokens. This contrast highlights the need for such a subagent. 

\section{Methodology}
In this section, we describe the design of the Execution Subagent and how it integrates into an existing coding agent, followed by our post-training pipeline used to produce Terminus-4B. Figure~\ref{train} shows a high-level overview of our approach. We begin by defining what a subagent is:

\begin{definition}[Subagent]
A secondary LLM-based agentic loop that is invoked by the parent (or the ``main'') agent to handle a specialized sub-task. Unlike the main agent, which solves a broad set of tasks, the subagent addresses a narrower, often simpler, set of tasks. Similar to the main agent, the subagent has its own system prompt, context window, and tool set that it uses to achieve the goal delegated to it by the main agent.
\end{definition}

\subsection{The Execution Subagent}
\textit{Execution Subagent} is one such subagent, which is specialized to sequentially generate and execute Terminal commands and return structured summaries of the results to the main agent. It is exposed to the main agent as a simple tool, which takes only the following two parameters:
\begin{itemize}
   \item \textbf{Query} (Required): A natural language description of the execution task i.e. what commands to run and what information to report back. For example: \textit{``Run the test suite and report which tests fail with their error messages.''}
   \item \textbf{Description} (Required): A short summary shown to the user in the UI while the subagent is executing.
\end{itemize}

The subagent returns a structured response containing a summary of each command that was run, its outcome and any relevant outputs (error messages, test counts, build status) delimited in XML-style (\texttt{<final\_answer>}) tags. This simple query-answer interface shields the main agent from the verbosity of the raw terminal output. The main agent can flexibly describe what it wants to be run and returned to it and the subagent handles how to get there.

\subsubsection{Tools \& Turn Limit}
The Execution Subagent has access to a single tool: \texttt{Terminal}. This tool takes a shell command, an execution mode (sync or async) and a timeout in milliseconds, and returns the shell command's output truncated to 60KB. This is the exact same tool that is given to the main agent for running terminal commands. Further, we also restrict the subagent to call this tool only once per turn i.e. no parallel tool calling, use sync mode with a time and be bounded by a configurable turn limit, which defaults to 10. If we reach the turn limit without the subagent exiting on its own, we inject a user message ( \textit{``OK, your allotted iterations are finished. Show the \texttt{<final\_answer>}.''}) coaxing a final answer from the subagent.

These constraints serve to simplify the design and keep the subagent focused and predictable. The subagent isn't given tools to read files, edit code or any other tools and can only run Terminal commands. The narrow scope is what makes this task amenable to a smaller finetuned model.

\subsubsection{System Prompt}
Figure~\ref{exec_system_prompt} shows the system prompt used for the Execution Subagent. The prompt instructs the model to: 1) run Terminal commands to accomplish the delegated tasks, adapting the commands as necessary, 2) follow specific rules for \texttt{Terminal} tool usage i.e. using sync mode, set explicit timeouts, no parallel calls, and 3) return a \texttt{<final\_answer>} containing a per-command summary with the command run and a concise description of the result. The prompt also includes an example showing a 2-step interaction (a failed \texttt{make} command followed by a successful \texttt{cmake \&\& make}) and the corresponding \texttt{<final\_answer>} formatted out. This example anchors the model's understanding of the expected behavior and output structure.

\subsubsection{Main Agent Integration}
The Execution Subagent is registered as one of the tools available to the main agent, along with tools like \texttt{ReadFile}, \texttt{Grep}, \texttt{Edit}, etc. We also augment the main agent's system prompt (Figure~\ref{main_prompt_diff}) to describe when / how to invoke the Execution Subagent.

We also add the ability to independently configure the model powering the subagent. The default configuration uses the same frontier model as the main agent (e.g., Claude Sonnet 4.6, GPT-5.3-Codex, etc.). While this produces high-quality trajectories and final responses, it is incredibly wasteful. The subagent's task of running commands and summarizing outputs is verbose and token-heavy. However, it is narrow well-structured compared to the main agent's open-ended planning and code editing. Using a frontier model for this is overkill. In the next section, we describe how we train Terminus-4B, a Qwen3-4B model post-trained specifically for this task, to replace the frontier model in the subagent while preserving, and in some cases even improving the overall agent performance.

% allowing a smaller or cheaper model to be used without affecting the main agent's model selection. This is the key property we exploit: by replacing the frontier model in the subagent with Terminus-4B, we reduce cost while preserving the agent performance.

\begin{figure}
    \centering
    \includegraphics[width=0.99\linewidth]{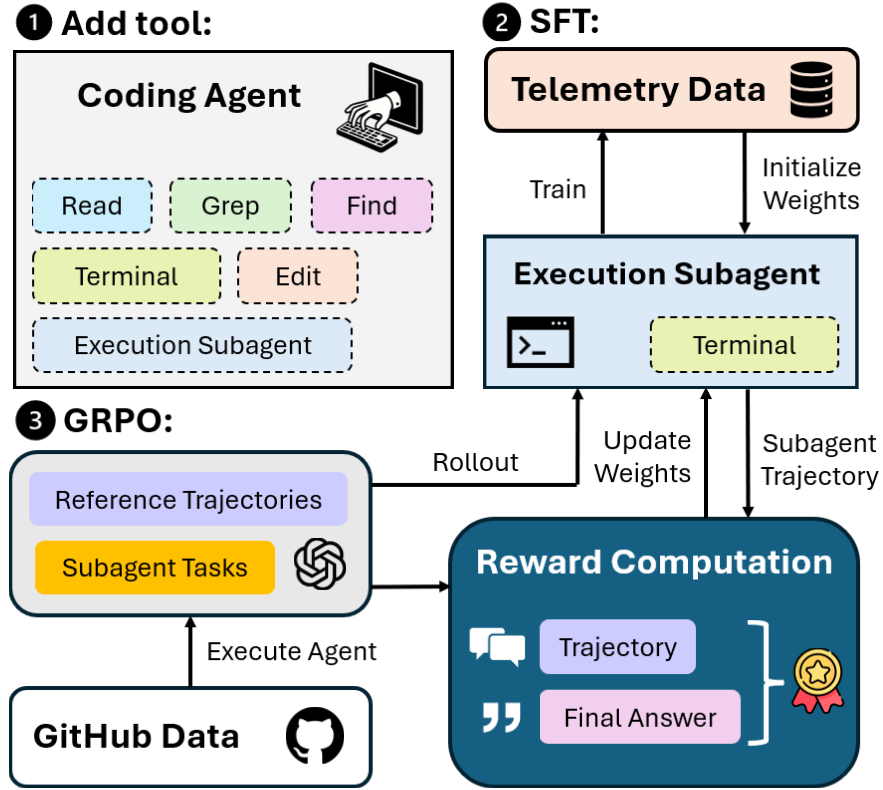}
    \caption{Terminus-4B training pipeline. \ding{182} We first integrate the Execution Subagent as a tool into the main agent framework. \ding{183} Then we do SFT where the base Qwen3-4B model is fine-tuned on expert trajectories from telemetry. \ding{184} We do further RL training with GRPO on examples from GitHub, where rrollouts are graded by a rubric-based LLM-as-judge reward against reference trajectories generated by Frontier LLM.}
    \label{train}
\end{figure}

\begin{figure}[t]
\centering
\fbox{%
\begin{minipage}{0.92\columnwidth}
\scriptsize
\ttfamily
\setlength{\parskip}{1pt}
\noindent You are an AI coding research assistant that runs terminal commands to perform a small execution-focused
task.\\[3pt]
You will be given a task description and potentially some commands to run, but you can adapt them as necessary.\\[3pt]
\textbf{Rules for run\_in\_terminal:}\\
- Always use mode="sync".\\
- Include "timeout" in ms (30000 short, 120000 builds).\\
- Only call run\_in\_terminal once per turn.\\
- Auto-confirm prompts with --yes, -y, or \texttt{yes}.\\[3pt]
\textbf{Output Format (REQUIRED)}\\
Once finished, return ONLY a \textless final\_answer\textgreater{} tag with a compact summary of each command:\\[2pt]
\phantom{x}\textless final\_answer\textgreater\\
\phantom{x}Command: cmake . \&\& make\\
\phantom{x}Summary: Build unsuccessful. Excerpt: ...\\
\phantom{x}\textless/final\_answer\textgreater\\[3pt]
\textbf{Tools Available}\\
- run\_in\_terminal: Execute a shell command
\end{minipage}%
}
\caption{System prompt for the Execution Subagent. The subagent is constrained to a single tool (\texttt{Terminal}), can only run sync commands and must return a structured \texttt{\textless final\_answer\textgreater} summary.}
\label{exec_system_prompt}
\end{figure}

\begin{figure}[t]
\centering
\fbox{%
\begin{minipage}{0.92\columnwidth}
\scriptsize
\ttfamily
\setlength{\parskip}{1pt}
\noindent You are a highly sophisticated automated coding agent with expert-level knowledge across many programming languages
and frameworks...\\[4pt]
\noindent The custom tools (Grep, FileSearch, ReadFile, ListDir) have been optimized for the agent surface.
Default to these over terminal commands...\\[5pt]
\colorbox{green!20}{\parbox{0.97\columnwidth}{\textbf{== Using ExecutionSubagent ==}\\[2pt]
For most execution tasks and terminal commands, use \texttt{ExecutionSubagent} to run commands and get relevant portions of
the output instead of using \texttt{Terminal}. Use \texttt{Terminal} in rare cases when you want the entire
output of a single command without truncation.\\[3pt]
Don't call \texttt{ExecutionSubagent} multiple times in parallel. Instead, invoke one subagent and wait for its response
before running the next command.}}\\[5pt]
\noindent When invoking a tool that takes a file path, always use the absolute file path...
\end{minipage}%
}
\caption{Instructions added to the main agent system prompt shown in green. We instruct the main agent to delegate terminal execution
to the Execution Subagent.}
\label{main_prompt_diff}
\end{figure}

\subsection{Post-Training Pipeline}
To train Terminus-4B, we conduct a two-stage post-training pipeline shown in Figure~\ref{train}. We first do supervised finetuning (SFT) on internal telemetry data consisting of expert trajectories, followed by reinforcement learning (RL) with Group Relative Policy Optimization (GRPO) on a curated dataset of GitHub issues and reward comparing rollouts to expert trajectories generated with a frontier LLM.

\subsubsection{Data Collection}
\label{data}
We construct a training dataset of $\sim$3200 execution tasks spanning multiple programming and projects on GitHub. We start with a pool of $\sim$10k instances across 2{,}144 repositories and 5 languages (TypeScript, C\#, Java, JavaScript, Python), each of which we confirmed to be a buildable repo. For each instance, the repository is checked out at the commit prior to the developer fix inside inside a Docker container with the project's dependencies pre-installed.

We then run our agent with a frontier LLM as the main agent on a subset of these instances (~2k) with the Execution Subagent enabled. During each run, the main agent is prompted to use the Execution Subagent and organically decides to invoke the subagent at various points in the trajectory. We extract each subagent invocation from the resulting trajectories, capturing 1) the \textit{query} issued by the main agent, 2) the complete subagent trajectory with all the \texttt{Terminal} calls, command outputs, and exit codes, the \texttt{<final\_answer>} returned to the main agent and 3) the state of the repository right before the subagent was called. The need for each of these fields will become clear as we further discuss the training pipeline.

\begin{table}[t]
\centering
\caption{Breakdown of collected tasks based on Language.}
\label{dataset_languages}
\scriptsize
\begin{tabular}{lr}
\toprule
\textbf{Language} & \textbf{Value} \\
\midrule
TypeScript & 1{,}011 (36.5\%) \\
C\# & 770 (27.8\%) \\
Java & 725 (26.2\%) \\
JavaScript & 196 (7.1\%) \\
Python & 65 (2.3\%) \\
\midrule
\end{tabular}
\end{table}

We found the resulting trajectories to contain 3{,}009 unique Execution Subagent invocations across 730 repositories. We take these trajectories as our gold-standard \textit{reference trajectories}. These trajectories will later be used to grade our rollouts during RL. Table~\ref{dataset_languages} shows the breakdown of the collected problems in terms of languages. We can see that the dataset is dominated by TypeScript and C\# due to our dataset's composition, but spans 720 repositories across 5 languages with diverse build systems using (\texttt{npm}, \texttt{dotnet}, \texttt{mvn}, \texttt{gradle}, \texttt{pip}), etc. Table~\ref{dataset_task_types} shows the kinds of tasks (classified using a simple prompt) represented in the data. We can see that the majority of tasks are test execution, error diagnosis or build tasks. This makes sense because usually running test requires building the project. Furthermore, compilation errors or failures to build the project typically requires error diagnosis.

\begin{table}[t]
\centering
\caption{Task Types represented in the collected set of execution tasks.}
\label{dataset_task_types}
\scriptsize
\begin{tabular}{lr}
\toprule
\textbf{Task Type (non-exclusive)} & \textbf{Value} \\
\midrule
Test Execution & 2{,}692 (97.3\%) \\
Build / Compile & 969 (35.0\%) \\
Error Diagnosis & 2{,}166 (78.3\%) \\
Dependency & 106 (3.8\%) \\
\bottomrule
\end{tabular}
\end{table}

\subsubsection{Phase I: Supervised Finetuning}
Before we do RL, we bootstrap the base Qwen3-4B model with supervised finetuning on expert trajectories collected from our internal usage telemetry. When the Execution Subagent is run with frontier model in production, each invocation produces a complete multi-turn trajectory with a system prompt, user query, tool calls with terminal outputs and the final \texttt(<final\_answer>). We extract these trajectories to use as SFT data. To prevent data leakage, we also ensure the SFT and RL datasets are drawn from disjoint repo sets.

We apply standard language modeling loss with loss masking i.e. gradients are only computed for assistant tokens, which equates to tool calls and the final answer and not on the system prompt, user messages and tool outputs:

\begin{equation}
\mathcal{L}_{\text{SFT}} = -\sum_{t \in \mathcal{A}} \log p_\theta(x_t \mid x_{<t})
\label{eq:sft}
\end{equation}

\noindent where $\mathcal{A}$ denotes the set of positions in the token sequence with assistant responses that get through masking. We do lightweight training for only two epochs. This stage teaches the model the mechanics of the task, including how to use the Terminal tool, %interpreting outputs, 
and writing a useful final response. This provides a strong initialization for our RL training. %We refer to this checkpoint as Terminus-SFT.

\subsubsection{Phase II: Reinforcement Learning}
Starting with the SFT checkpoint, we performr on-policy RL with Group Relative Policy Optimization (GRPO)~\cite{grpo} on the 2.7k execution tasks collected in Section~\ref{data}. While SFT teaches the model the overall task shape, it doesn't necessarily optimize for the key qualities of an execution subagent such as accurate final answer with sufficient detail. RL with carefully designed reward allows us to directly shape such behaviors. However, we must first discuss how we do rollouts at scale.

\paragraph{Subagent Rollout Framework.}
One key challenge in training agents with RL is that each rollout requires end to end execution and must begin at the same starting point (i.e. same initial prompt, problem, repo state, etc.). While this is simpler to do if we were training the main agent, it's slightly trickier to achieve this kind of consistency between subagent rollouts.

We address this by decoupling the main agent from the subagent during training. Each training instance comes with a pre-generated query (i.e. the input the Execution Subagent) and repo state collected during the data collection step (Section~\ref{data}). During rollouts, we use a lightweight pass-through model (Qwen3-4B Instruct-2507) as the main agent, which has been configured to just have a single tool (\texttt{Execution Subagent}) and a 1-turn limit. This ensures the main agent deterministically forwards the pre-generated query to the subagent on its first and only turn. We wrap each query into a simple instruction as follows:
\begin{center}
\small
\fbox{\parbox{0.78\columnwidth}{\ttfamily\scriptsize
Call Execution Subagent with this query. You MUST\\
use Execution Subagent ONLY with this EXACT query.\\[4pt]
<query>\\
Run \textasciigrave dotnet build PdfSharpCore.sln\textasciigrave{} in /testbed to check\\
for compilation errors, then run \textasciigrave dotnet test\\
PdfSharpCore.Test.csproj\textasciigrave{} and report any test failures.\\
</query>
}}
\end{center}

It should be noted that while we ensure deterministic behavior of the main agent, the subagent's behavior is then entirely determined by the policy model being trained and is able to run for a full trajectory, after which the main agent trajectory concludes. A key advantage of decoupling the main agent from subagent like this is that it allows us to eliminate the cost of using a Frontier LLM as the main agent, helping us achieve fast and inexpensive rollouts.

Each of these rollouts executes inside a Docker container that contains our agent with the necessary modifications, the target repo checked out with the correct commit and the dependencies needed to build the project and run the agent. Before rollout, we also apply a code patch to ensure the repo state matches the state before Execution Subagent was called in reference trajectory. Depending on when the Execution Subagent was called in the reference trajectory, the patch may or may not be empty. Finally, we use Azure Container Apps (ACA) to orchestrate parallel rollouts.

\paragraph{Reward Design.}
Designing an effective reward is critical for RL to improve on top of SFT. Grading the quality of execution subagent rollouts is a difficult task as many things can go wrong. For instance, the subagent may execute correct commands, but produce a vague summary or write a detailed summary that has hallucinated information. Further, it's hard to compare raw trajectories directly as two valid runs may have completely differnt commands that produce different outputs.

We design a rubric-based LLM-as-judge reward that breaks our reward into two subcomponents based on our two main concerns: \textit{did the subagent execute the right commands?} and \textit{how useful was the final answer to the main agent's workflow?}. The second component is especially important because the final summary is the onlt window the main agent has into what the subagent did.

\begin{figure}[t]
\centering
\fbox{%
\begin{minipage}{0.92\columnwidth}
\scriptsize
\ttfamily
\setlength{\parskip}{1pt}
\noindent\textbf{System:} You are a helpful assistant that creates execution plans summarizing what an AI agent did when running terminal commands. Your plans should follow this template:\\[3pt]
\textbf{EXECUTION PLAN}\\[2pt]
\textbf{Task Outcome:} [success / failure / partial. One sentence.]\\[2pt]
\textbf{Commands Executed:} [For each command: what was run, why, and what happened. Include exit codes.]\\[2pt]
\textbf{Key Findings:} [Error messages, version numbers, test counts, build artifacts, etc.]\\[2pt]
\textbf{Error Recovery:} [How were failures handled? What adaptations were made?]\\[2pt]
\textbf{Final State:} [What was accomplished?]\\[6pt]
\textbf{User:}\\
\#\# Task Query\\
\{query\}\\[2pt]
\#\# Trajectory Data\\
\#\#\# Command 1: \textasciigrave cmake . \&\& make\textasciigrave\\
Exit Code: 0\\
Output: \textit{(truncated head + tail)}\\[2pt]
\#\#\# Command 2: \textasciigrave make test\textasciigrave\\
Exit Code: 1\\
Output: \textit{(truncated head + tail)}\\[2pt]
\#\# Final Answer Returned to Parent Agent\\
\{final\_answer\}\\[2pt]
Create a detailed execution plan following the template.
\end{minipage}%
}
\caption{Prompt used with a frontier LLM to generate Execution Plans from raw trajectories. Command outputs are truncated to the first and last 500 characters to keep the prompt concise while preserving key information (errors typically appear at the end, setup info at the start). We use this to convert both the reference and rollout trajectories to their corresponding Execution Plans.}
\label{execution_plan_prompt}
\end{figure}

To compare what the subagent did with the reference, one could pass the entire trajectories (rollout and ref) to the LLM, but that would be very expensive. We would need to process the reference trajectory once for every rollout's reward computation i.e. N times the cost. Rather than comparing verbose raw trajectories, we first condense each rollout's trajectory into a structured summary we call an \textit{Execution Plan} using the prompt shown in Figure~\ref{execution_plan_prompt}. We give a frontier LLM a condensed version of the trajectory where we include all the commands
The same extraction was applied to the reference trajectories during data collection, producing reference execution plans as ground truth. This intermediate representation normalizes trajectories into a comparable format regardless of the specific commands used. We then use a powerful frontier LLM for grading the candidate plan against the reference plan across 14 rubric dimensions across three groups:
\begin{itemize}
    \item \textbf{Execution Quality}: We ask the grader to evaluate the candidate plan along 7 dimensions: command correctness, error handling, outcome accuracy, key information extraction, completeness, efficiency and actionability. We average out these scores to get $\bar{s}_{\text{pos}}$.
    \item \textbf{Failure Modes}: We score 4 failure mode dimensions: hallucinated results, missed errors, wrong diagnosis and unnecessary commands. Averaging these gives us $\bar{s}_{\text{pit}}$.
    \item \textbf{Final Answer Quality}: We score 3 dimensions assessing the quality of the final answer returned to the main agent: detail level, factual accuracy, and informativeness. This group targets the \texttt{<final\_answer>} specifically, since it is the \textit{only} output the main agent sees. Averaging the scores gives us $\bar{s}_{\text{fa}}$.
\end{itemize}

The final reward blends the scores for execution quality with final answer quality:
\begin{equation}
r = (1 - \alpha)(\bar{s}_{\text{pos}} -
\bar{s}_{\text{pit}}) + \alpha \cdot
\bar{s}_{\text{fa}}
\label{eq:reward}
\end{equation}
where $\bar{s}_{\text{pos}}$, $\bar{s}_{\text{pit}}$, and
$\bar{s}_{\text{fa}}$ are the means of the positive, pitfall, and final answer dimension scores respectively. We use an $\alpha=0.5$.

Before rubric gading, we also apply some hard penalties for degenerate rollouts. Trajectories exceeding 30K tokens receive $r = -100$. The ones missing \texttt{<final\_answer>} tags receive $r = -100$, and
rollouts with no commands receive $r = -50$. We also discard prompt groups where the reward standard deviation $\sigma_G < 0.01$, as near-identical scores across all $G$ rollouts would produce negligible gradient signal for GRPO to learn.

\begin{figure*}[t]
   \centering
   \includegraphics[width=0.87\linewidth]{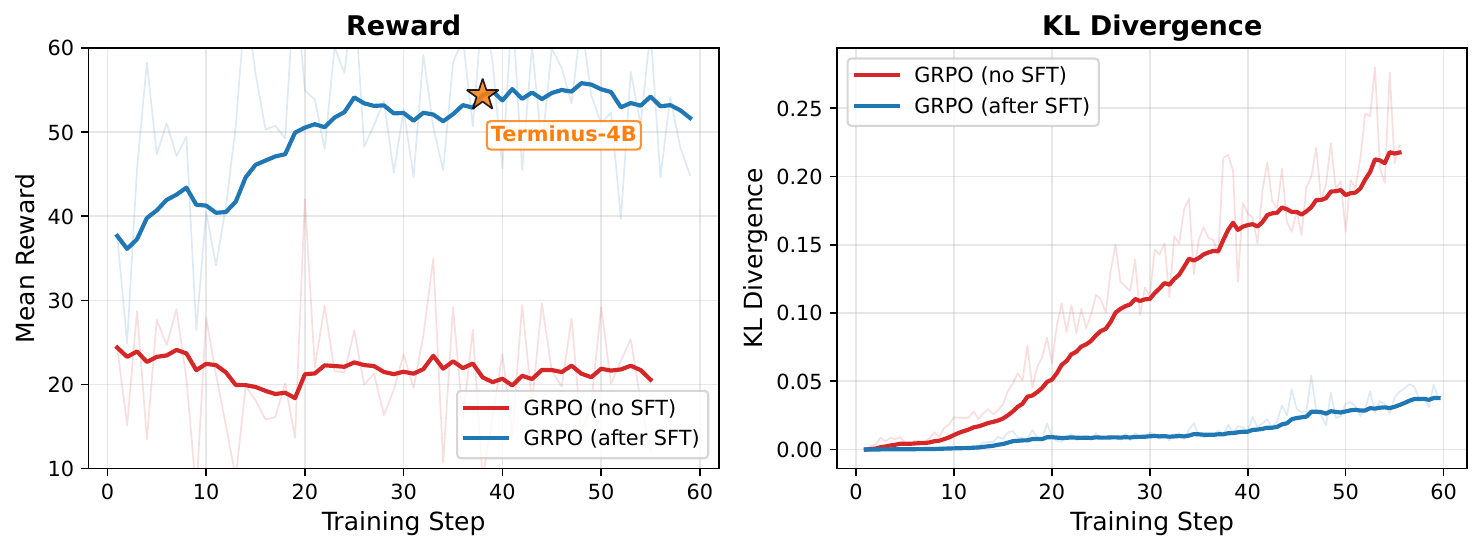}
   \caption{Training plots for GRPO with and without SFT initialization. SFT starts RL training off with a strong prior that enables higher reward and learning. Looking at the KL plot, we can see that the run stays close to reference (SFT checkpoint). We can clearly see the gain from starting from the SFT checkpoint.}
   \label{training_curves}
\end{figure*}

\section{Experimental Setup}
\subsection{Training Setup}
Across all our training runs, we used a single node equipped with 8 NVIDIA A100/H100 GPUs for training. For inference we deploy the model on Fireworks platform~\cite{fireworks}.

\subsubsection{SFT Details} We first perform full SFT on the Qwen3-4B-Instruct-2507 base model with Slime~\cite{slime} using telemetry data with the subagent. We train for 2 epochs with the Adam optimizer ($\beta_1{=}0.9$, $\beta_2{=}0.95$), a peak learning rate of $2 \times 10^{-5}$ with cosine decay to $2 \times 10^{-6}$ and 5\% linear warmup. We use a global batch size of 32, weight decay of 0.01, and dynamic batching with a maximum of 8{,}192 tokens per GPU. %Loss is computed only on assistant tokens.

\subsubsection{RL Details} Starting from the SFT checkpoint, we apply on-policy RL using GRPO with asymmetric clipping~\cite{dapo} to reduce premature collapse. For each prompt $x$, we sample $G=8$ rollouts $\{a_i\}_{i=1}^{G} \sim \pi_{\theta_{\text{old}}}(\cdot \mid x)$. We optimize the following objective:

{\fontsize{6.8pt}{10pt}\selectfont
\begin{equation}
\begin{split}
   J(\theta) = \frac{1}{G} \sum_{i=1}^{G}
   \min\!\Big(
     r_i(\theta)\,\hat{A}_i,\; \text{clip}\big(r_i(\theta),
       1{-}\epsilon_l,
       1{+}\epsilon_h\big)\hat{A}_i
   \Big) - \beta\, D_{\text{KL}}\!\big[\pi_\theta \| \pi_{\text{SFT}}\big]
   \end{split}
\end{equation}
}

\noindent where $r_i(\theta) = \frac{\pi_\theta(a_i \mid x)}{\pi_{\theta_{\text{old}}}(a_i \mid x)}$ is the importance ratio, $\hat{A}_i$ is the group normalized advantage and we use asymmetric clip thresholds of $\epsilon_{\text{low}} = 0.20$, $\epsilon_{\text{high}} = 0.28$ to allow the policy more room to increase high reward probabilities. The KL penalty we use is $\beta=0.02$, which is computed using hte SFT checkpoint as the reference model. This is to prevent catastrophic drift from the reference and base model behavior. We train with a constant learning rate of $10^{-6}$ and gradient clipping at $1.0$. We use a rollout batch size of 16 prompts (each with $G{=}8$ samples) i.e. yielding 128 rollouts per step with a global training batch size of 64. For reward computation, execution plans are generated and graded with a frontier LLM.

\subsection{Evaluation Setup}
In this section, we lay out the offline evaluations we conducted on our technique, the benchmarks used and metrics we measured.
\subsubsection{Benchmarks}
We use evaluate our approach on two benchmarks:
\begin{itemize}
    \item \textbf{SWE-Bench Pro}: This is a benchmark~\cite{swepro} containing problems from a diverse set of examples and is explicitly designed to solve some of the problems SWE-Bench~\cite{swebenchverified} faced. These span multiple languages and domains. We run $\sim$731 examples from the full set.
    \item \textbf{TerminalBench 2.0}: This is a benchmark~\cite{terminalbench} consisting of 89 carefully curated tasks based on real workflows, which have been shown to be challenging for Frontier LLMs.
    \item \textbf{SWE-Bench C\#}: This is our internal benchmark of 150 GitHub issues collected from C\# repositories in SWE-Bench style of collection. Since solving the problems in this benchmark requires lots of project building / running tests, this is a natural candidate for evaluating our approach.
\end{itemize}

\subsubsection{Ablations}
Given these benchmarks, we conduct a set of ablation experiments designed to test two things: 1) end-to-end agent performance with our approach, i.e. our approach doesn't compromise the agent's ability to solve problems or negatively impact token consumption, etc. 2) subagent output quality, i.e. does the main agent actually find the subagent output useful?

For our experiments, we use the following subagent configurations ranging from no subagent to various models we want to measure performance of:
\begin{itemize}
    \item \textbf{Opus / Sonnet}: We use Frontier LLMs Opus / Sonnet as subagents. These are meant to represent the upper bound in terms of capability.
    \item \textbf{Vanilla-4B}: We use the base Qwen3-4B-Instruct-2507 model as the subagent.
    \item \textbf{SFT-4B}: We use the checkpoint following SFT as the subagent.
    \item \textbf{Terminus-4B}: We use the peak checkpoint after RL as the model behind the subagent.
\end{itemize}

We combine these subagent configurations with 3 frontier LLMs as main agents: GPT-5.3-Codex, Claude Sonnet 4.6, Claude Opus 4.6, in order of increasing capability.

\subsubsection{Tool Configurations}
We ablate over different tool configurations to isolate the impact of the subagent on the agent's performance:
\begin{itemize}
    \item \textbf{Terminal Tool Only (No Subagent)}: This is our baseline where the Execution Subagent is disabled entirely. We also leave the system prompt of the main agent unmodified.
    \item \textbf{Subagent + Terminal Tool}: The Execution Subagent is enabled along with the Terminal tool. The main agent has the choice to delegate terminal work to the subagent or can run directly using Terminal.
    \item \textbf{Subagent Only (No Terminal)}: The Execution Subagent is the only terminal execution tool available to the main agent.
\end{itemize}

\begin{figure}[t]
\centering
\fbox{%
\begin{minipage}{0.92\columnwidth}
\scriptsize
\ttfamily
\setlength{\parskip}{1pt}
\noindent\textbf{System:} You are a strict evaluator assessing the
quality of an execution subagent's response. The main agent only
sees the subagent's final response — it has no access to
intermediate commands or raw output.\\[3pt]
Judge the response on these dimensions (0.0--1.0):\\[2pt]
\textbf{task\_completion}: Did the subagent fully accomplish what
was asked?\\
\textbf{factual\_accuracy}: Is the response factually grounded
and verifiable? (exit codes, paths, exact errors)\\
\textbf{informativeness}: Enough detail to proceed without
re-running commands?\\
\textbf{relevance}: Focused on the query with no extraneous
content?\\
\textbf{actionability}: Can the main agent determine clear next
steps?\\[4pt]
\noindent\textbf{User:}\\
\#\# Main Agent Context\\
System prompt (excerpt): \{system\_prompt\}\\
Problem statement: \{problem\_statement\}\\
Trajectory so far: \{trajectory\} \\[2pt]
\#\# Subagent Query\\
\{subagent\_query\}\\[2pt]
\#\# Subagent Response\\
\{subagent\_response\}\\[2pt]
\#\# Subsequent Trajectory (\{N\} turns after)\\
\{trajectory\_after\}\\[2pt]
Judge the response using the subsequent trajectory to assess
whether the main agent effectively used it.
\end{minipage}%
}
\caption{A condensed version of the LLM-judge Score Prompt. The judge receives the main agent's full context before and after the subagent call, and scores the response along 5 dimensions. The subsequent N=5 turns in the trajectory after the subagent call reveals whether the response was useful in practice.}
\label{judge_prompt}
\end{figure}

\subsubsection{Metrics}
For all our configurations, we compute the following metrics using the evaluation harness outcome and agent trajectories:
\begin{itemize}
    \item \textbf{Resolution Rate (\%)}: The fraction of instances that pass the benchmark's evaluation.
    \item \textbf{Token Usage}: This is the average input and output main agent / subagent tokens consumed by the agent configuration.
    \item \textbf{Main Agent Terminal Calls}: This is the average direct calls to Terminal tool made by the main agent.
    \item \textbf{Subagent followed by Terminal Calls}: This the average number of cases per instance where the Terminal tool was called right after Execution Subagent. We expect a bulk of these cases to be the main agent having to redo the work done by the subagent due to the output not being very useful.
    \item \textbf{Final Answer Rate (\%)}: We measure the fraction of subagent calls that return a well-formed final answer delimited with \texttt(<final\_answer>) tags.
    \item \textbf{LLM-Judge Score}: To get a measure of the utility of the final answer from subagent, we compute an LLM score where the LLM is given the trajectory leading up to the subagent call, the final response and N=5 steps coming after the subagent call to see if the results were actually used. The LLM is prompted to score the final response along various dimensions. The prompt used is shown in Figure~\ref{judge_prompt}.
\end{itemize}

\section{Results}
\subsection{RL Training}
Figure~\ref{training_curves} shows how the reward and KL divergence tends over the course of our training for two configurations: GRPO applied directly to the base Qwen3-4B-Instruct-2507 model (i.e. no SFT) and GRPO starting from the SFT checkpoint.

Comparing the reward plots, we can see that the reward for the `No SFT' run plateaus at $\sim$20 and fails to improve compared to the baseline scores, seen initially in the run. However, the KL divergence rises rapidly to over 0.2, indicating the policy drifted significantly from reference without any meaningful learning. We believe this is because the base model lacks a fundamental understanding of this task and things like output format and doesn't manage to stumble on those behaviors during training to provide GRPO the gradient signal needed to enforce such behaviors. 

In stark contrast, we can see that the RL run starting with the SFT checkpoint starts at a much higher reward of $\sim$37 and steadily climbs 50+ reward, while the KL remains close to the SFT checkpoint (within 0.05). Intuitively, this is because SFT is able to provide the model with a good starting knowledge of the mechanics of the task i.e. what's the desired output format, the Terminal tool usage, etc. These results confirm that SFT is not merely helpful, but essential to the training. We call the resulting model from our RL training the \textbf{Terminus-4B} model. We show the utility of the RL training itself via our ablations.

\subsection{Benchmark Ablations}
In this subsection, we talk about the different ablations we conduct on the benchmarks in our evaluation setup.

%\begin{table*}[t]
%\centering
%\caption{Results of running various models as the Execution Subagent over SWE-Bench Pro, using %Claude Opus 4.6 as the main agent model. We show the tokens consumed by frontier LLMs for each %run, as well as SLM tokens tokens consumed. We can see that with both SFT-4B and Terminus-4B, %expensive frontier token usage drops while overall performance is maintained.}
%\label{swepro}
%\scriptsize
%\begin{tabular}{lccccccc}
%\toprule
%\textbf{Subagent Configuration} & \textbf{Resolve \%} & \textbf{Main Agent Tokens} & %\textbf{Subagent Tokens} & \textbf{Frontier LLM Tokens} & \textbf{Main Agent Terminal} & %\textbf{Subagent$\to$Terminal} &
%\textbf{Final Answer \%} \\
%\midrule
%No Subagent  & 30.0 & 836k & -  & 836k  & 3.8 & -  & -   \\
%Opus         & 32.0 & 729k & 19k  & 748k  & \textbf{0.5} & \textbf{0.04} & \textbf{100.0} \\
%Sonnet       & \textbf{32.6} & 756k & 19k  & 774k  & 0.6 & 0.06 & \textbf{100.0} \\
%Vanilla-4B   & 30.4 & 840k & 6k   & 840k  & 1.9 & 0.27 & 99.0  \\
%SFT-4B       & 32.4 & \textbf{727k} & 17k  & \textbf{727k}  & 1.1 & 0.17 & 98.5  \\
%Terminus-4B       & 31.5 & 730k & 18k  & 730k  & 1.0 & 0.14 & 97.9  \\
%\bottomrule
%\end{tabular}
%\end{table*}

\begin{table*}[t]
\centering
\caption{Results of running various models as the Execution Subagent over SWE-Bench Pro, using Claude Opus 4.6 as the main agent model. We show the tokens consumed by frontier LLMs for each run, as well as SLM tokens consumed. We can see that with both SFT-4B and Terminus-4B, expensive frontier token usage drops while overall performance is maintained.}
\label{swepro}
\scriptsize
\begin{tabular}{llllllll}
\toprule
\textbf{Subagent Config.} & \textbf{Resolve \%} & \textbf{Main Agent Tokens} & \textbf{Subagent Tokens} & \textbf{Frontier LLM Tokens} & \textbf{Main Agent Terminal} & \textbf{Subagent$\to$Terminal} & \textbf{Final Answer \%} \\
\midrule
No Subagent  & 30.0 & 836k & -  & 836k  & 3.8 & -  & -   \\
Opus         & 32.0 & 729k (\impact{-12.8\%}{green}) & 19k  & 748k (\impact{-10.5\%}{green})  & 0.5 (\impact{-86.8\%}{green}) & 0.04 & 100.0 \\
Sonnet       & 32.6 & 756k (\impact{-9.6\%}{green}) & 19k  & 774k (\impact{-7.4\%}{green})  & 0.6 (\impact{-84.2\%}{green}) & 0.06 & 100.0 \\
Vanilla-4B   & 30.4 & 840k (\impact{+0.5\%}{red}) & 6k   & 840k (\impact{+0.5\%}{red})  & 1.9 (\impact{-50.0\%}{green}) & 0.27 & 99.0  \\
SFT-4B       & 32.4 & 727k (\impact{-13.0\%}{green}) & 17k  & 727k (\impact{-13.0\%}{green})  & 1.1 (\impact{-71.1\%}{green}) & 0.17 & 98.5  \\
Terminus-4B  & 31.5 & 730k (\impact{-12.7\%}{green}) & 18k  & 730k (\impact{-12.7\%}{green})  & 1.0 (\impact{-73.7\%}{green}) & 0.14 & 97.9  \\
\bottomrule
\end{tabular}
\end{table*}

\begin{table}[t]
\centering
\caption{Resolution rates (\%) and subagent call rates of different subagent and main agent model combinations over SWE-Bench C\#.}
\label{swebcs_resolve}
\scriptsize
\begin{tabular}{lcccccc}
\toprule
& \multicolumn{2}{c}{\textbf{Claude Opus 4.6}} & \multicolumn{2}{c}{\textbf{Claude Sonnet 4.5}} & \multicolumn{2}{c}{\textbf{GPT-5.3-Codex}} \\
\textbf{Subagent} & \tiny Resolve & \tiny Call \% & \tiny Resolve & \tiny Call \% & \tiny Resolve & \tiny Call \% \\
\midrule
No Subagent   & 46.7 & - & 47.3 & - & 38.0 & - \\
Opus          & 46.0 & 94.6 & 46.7 & 75.3 & 40.3 & 89.3 \\
Sonnet        & 46.0 & 93.3 & 44.7 & 72.0 & 40.7 & 90.7 \\
Vanilla-4B    & 44.0 & 91.0 & 44.0 & 74.3 & 36.9 & 87.1 \\
SFT-4B        & 47.0 & 95.3 & 44.0 & 73.2 & 42.0 & 89.9 \\
Terminus-4B   & 46.7 & 89.3 & 46.7 & 74.7 & 38.0 & 85.3 \\
\bottomrule
\end{tabular}
\end{table}

\begin{table*}[t]
\centering
\caption{Results of running various models as Execution Subagent with different main agent models for SWE-Bench C\#. We show the breakdown of token usage as well as behavioral metrics. We show percentage changes relative to the No Subagent baseline. For Frontier tokens, we include main agent tokens plus subagent tokens when the Execution Subagent is a frontier model. SLM tokens are consumed by only the 4B subagents.}
\label{swebcs_metrics}
\scriptsize
\begin{tabular}{l|llll|llll|llll}
\toprule
\textbf{Subagent} & \multicolumn{4}{c|}{\textbf{Claude Opus 4.6}} & \multicolumn{4}{c|}{\textbf{Claude Sonnet 4.5}} & \multicolumn{4}{c}{\textbf{GPT-5.3 Codex}} \\
\midrule
\multicolumn{13}{c}{\textbf{Avg. Tokens}} \\
\midrule
& \tiny Main & \tiny Sub & \tiny Frontier & \tiny SLM & \tiny Main & \tiny Sub & \tiny Frontier & \tiny SLM & \tiny Main & \tiny Sub & \tiny Frontier & \tiny
SLM \\
No Subagent   & 1{,}010k & - & 1{,}010k & - & 1{,}496k & - & 1{,}496k & - & 1{,}036k & - & 1{,}036k & - \\
Opus          & 710k (\impact{-29.7\%}{green}) & 25k & 734k & - & 1{,}125k (\impact{-24.8\%}{green}) & 26k & 1{,}151k & - & 764k (\impact{-26.3\%}{green}) & 25k & 789k & - \\
Sonnet        & 757k (\impact{-25.0\%}{green}) & 27k & 784k & - & 1{,}074k (\impact{-28.2\%}{green}) & 28k & 1{,}102k & - & 758k (\impact{-26.8\%}{green}) & 30k & 789k & - \\
Vanilla-4B    & 802k (\impact{-20.6\%}{green}) & 11k & 802k & 11k & 1{,}460k (\impact{-2.4\%}{green}) & 9k & 1{,}460k & 9k & 902k (\impact{-12.9\%}{green}) & 12k & 902k & 12k \\
SFT-4B        & 833k (\impact{-17.5\%}{green}) & 23k & 833k & 23k & 1{,}109k (\impact{-25.9\%}{green}) & 24k & 1{,}109k & 24k & 726k (\impact{-29.9\%}{green}) & 36k & 726k & 36k \\
Terminus-4B   & 693k (\impact{-31.4\%}{green}) & 25k & 693k & 25k & 1{,}177k (\impact{-21.3\%}{green}) & 27k & 1{,}177k & 27k & 704k (\impact{-32.0\%}{green}) & 42k & 704k & 42k \\
\midrule
\multicolumn{13}{c}{\textbf{Main Agent Terminal Usage \& Subagent$\to$Terminal}} \\
\midrule
& \multicolumn{2}{c}{\tiny Main Terminal} & \multicolumn{2}{c|}{\tiny Sub$\to$Terminal} & \multicolumn{2}{c}{\tiny Main Terminal} & \multicolumn{2}{c|}{\tiny Sub$\to$Terminal} & \multicolumn{2}{c}{\tiny Main Terminal} & \multicolumn{2}{c}{\tiny Sub$\to$Terminal} \\
No Subagent   & \multicolumn{2}{c}{6.2} & \multicolumn{2}{c|}{-} & \multicolumn{2}{c}{6.3} & \multicolumn{2}{c|}{-} & \multicolumn{2}{c}{4.2} & \multicolumn{2}{c}{-} \\
Opus          & \multicolumn{2}{c}{0.7 (\impact{-88.7\%}{green})} & \multicolumn{2}{c|}{0.09} & \multicolumn{2}{c}{1.0 (\impact{-84.1\%}{green})} & \multicolumn{2}{c|}{0.06} & \multicolumn{2}{c}{0.0 (\impact{-100\%}{green})} & \multicolumn{2}{c}{0.00} \\
Sonnet        & \multicolumn{2}{c}{0.9 (\impact{-85.5\%}{green})} & \multicolumn{2}{c|}{0.13} & \multicolumn{2}{c}{1.1 (\impact{-82.5\%}{green})} & \multicolumn{2}{c|}{0.11} & \multicolumn{2}{c}{0.1 (\impact{-97.6\%}{green})} & \multicolumn{2}{c}{0.02} \\
Vanilla-4B    & \multicolumn{2}{c}{2.5 (\impact{-59.7\%}{green})} & \multicolumn{2}{c|}{0.39} & \multicolumn{2}{c}{3.6 (\impact{-42.9\%}{green})} & \multicolumn{2}{c|}{0.36} & \multicolumn{2}{c}{1.4 (\impact{-66.7\%}{green})} & \multicolumn{2}{c}{0.43} \\
SFT-4B        & \multicolumn{2}{c}{2.1 (\impact{-66.1\%}{green})} & \multicolumn{2}{c|}{0.31} & \multicolumn{2}{c}{2.1 (\impact{-66.7\%}{green})} & \multicolumn{2}{c|}{0.23} & \multicolumn{2}{c}{0.8 (\impact{-81.0\%}{green})} & \multicolumn{2}{c}{0.25} \\
Terminus-4B   & \multicolumn{2}{c}{1.7 (\impact{-72.6\%}{green})} & \multicolumn{2}{c|}{0.23} & \multicolumn{2}{c}{2.4 (\impact{-61.9\%}{green})} & \multicolumn{2}{c|}{0.17} & \multicolumn{2}{c}{0.9 (\impact{-78.6\%}{green})} & \multicolumn{2}{c}{0.19} \\
\bottomrule
\end{tabular}
\end{table*}

\subsubsection{Cross-Language Generalization (via SWE-Bench Pro)}
To test whether the Execution Subagent architecture and Terminus-4B generalize to different languages, we evaluate on SWE-Bench Pro~\cite{swepro}, which is a multilingual benchmark spanning Python, JavaScript, TypeScript, Java, Go, etc. We use Claude Opus 4.6 as the main agent model across all subagent configurations. Table~\ref{swepro} shows the results.

We see that in terms of token usage, all subagent configurations improve over the No Subagent configuration. We also see no impact to the agent's performance on the benchmark as all the Resolve \% numbers stay close to the baseline resolution rate of 30\%. This confirms that delegating terminal execution to a subagent is beneficial in complex coding tasks. 

Looking at the Frontier token usage, we see savings for both the SFT-4B and Terminus-4B models ($\sim$13\% or $\sim$110k tokens per instance for both), which is higher than the Frontier LLM token usage of Sonnet or Opus as the subagent. We can see a clear improvement over Vanilla Qwen3-4B model where the token usage actually increases (by $\sim$0.5\%) over the baseline as the main agent likely has to compensate for unhelpful responses from the subagent by having to run Terminal commands on its own, which we can see from the high Subagent$\to$Terminal of 0.27, which means that over a quarter of Execution Subagent calls using the Vanilla model are followed by a Terminal call by the main agent. This clearly shows that our training improves on the Vanilla model.

Looking more closely at the behavioral metrics, we see that the Main Agent's Terminal usage drops from an average call rate of 3.8 per instance to 1 for our RL-trained model, which is a $\sim$74\% reduction. This is an improvement over the Vanilla and SFT models (1.9 and 1.1 calls per instance, respectively). We also see a similar reduction in the Subagent$\to$Terminal for Terminus-4B, where only 14\% of subagent calls are followed by terminal usage from the main agent. While this is still far from Opus and Sonnet's performance (4\% and 6\%) on this metric, it's a big improvement from the Vanilla and SFT models. These results demonstrate that the execution behaviors learned during our post-training i.e. Terminal tool usage patterns, final answer generation with effective summarization of subagent context, transfer effectively across programming languages.

\subsubsection{Generalization Across Main Agent Models (via SWE-Bench C\#)}
To test whether the subagent and Terminus-4B work regardless of the Frontier LLM used as the main agent, we evaluate our approach on SWE-Bench C\# with three main agent models: Claude Opus 4.6, Sonnet 4.6 and GPT-5.3-Codex. Table~\ref{swebcs_resolve} shows the resolution rates and relative call rates of the subagent between different main agent models. Table~\ref{swebcs_metrics} shows the token usage and behavioral metrics across the different configurations.

Looking at the Table~\ref{swebcs_resolve}, we can see that the resolution rates remain quite stable across all subagent configurations, with Terminus-4B matching or being close to the No Subagent baseline for all main agent models. This confirms that the Subagent does not degrade the end-to-end performance of the agent regardless of the main agent model. The call rate shows how often different models choose to call the subagent. We see that Opus is more likely to call the Execution Subagent ($>$90\% of instances) with Sonnet being somewhat lower ($\sim$72 - 75\%).

In terms of token efficiency metrics in Table~\ref{swebcs_metrics}, we can see that Terminus-4B achieves the largest Main agent and Frontier LLM token reduction for Claude Opus 4.6 and GPT-5.3-Codex where the subagent usage remains high. We can clearly see that these reductions in token usages are higher than Frontier LLMs. For Sonnet as the main agent, the custom trained models perform comparably to Opus and Sonnet as subagents. The Vanilla Qwen3-4B model shows the weakest savings throughout, showing the importance of our post-training for this task.

Finally, the behavioral metrics also show a consistent pattern across all three main agents. Terminus-4B reduces the main agent's Terminal calls by 62-79\% compared to the No Subagent baseline. We also see the Subagent$\to$Terminal distrust signal drop due to our post-training. Terminus-4B achieves a substantial improvement over both Vanilla and SFT in terms of the distrust signal across all main agent models. This shows the robustness of the quality improvements from RL and that they transfer across different main agent models.

\begin{table*}[t]
\centering
\caption{Results on SWE-Bench C\# results after Terminal tool is removed from the main agent, using Claude Opus 4.6 as the main agent model. All terminal work must flow through the Execution Subagent, providing a cleanest comparison of subagent quality. We show percentage changes relative to the Opus subagent baseline.}
\label{swebcs_no_terminal}
\scriptsize
\begin{tabular}{lccclcclcc}
\toprule
\textbf{Subagent} & \textbf{Resolve \%} & \textbf{Main Agent Tokens} & \textbf{Subagent Tokens} & \textbf{Frontier LLM Tokens} & \textbf{SLM Tokens} & \textbf{Sub Calls} & \textbf{Sub$\to$Sub} & \textbf{Final Ans. \%} \\
\midrule
Opus         & 45.3 & 704k & 35k & 740k & - & 2.4 & 0.89 & 100.0 \\
Sonnet       & 48.0 & 632k & 33k & 664k (\impact{-10.3\%}{green}) & - & 2.3 & 0.76 (\impact{-14.6\%}{green}) & 99.4 \\
Vanilla-4B   & 44.7 & 810k & 18k & 810k (\impact{+9.5\%}{red}) & 18k & 3.1 & 1.51 (\impact{+69.7\%}{red}) & 99.4 \\
SFT-4B       & 45.3 & 590k & 32k & 590k (\impact{-20.3\%}{green}) & 32k & 2.5 & 1.07 (\impact{+20.2\%}{red}) & 98.7 \\
Terminus-4B  & 45.9 & 605k & 34k & 605k (\impact{-18.2\%}{green}) & 34k & 2.4 & 0.89 (\impact{0.0\%}{green}) & 98.9 \\
\bottomrule
\end{tabular}
\end{table*}

\subsubsection{Removing Terminal Tool}
To isolate the quality of the subagent from the main agent's ability to compensate for poor responses by running Terminal tool on its own, we run an ablation where the Terminal tool is removed entirely from the main agent. In this configuration, all terminal execution must flow through the Execution Subagent. This provides a clean comparison of the subagent quality and stress test across different subagent models. We use Claude Opus 4.6 as the main agent. Further, rather than showing Subagent$\to$Terminal metric in our previous evaluations, we show Subagent$\to$Subagent as a way to measure how often the main agent needs to repeat subagent calls. Table~\ref{swebcs_no_terminal} shows the results of this ablation.

Once again the resolution rates remain comparable across all configurations (44-48\% range), confirming that even without Terminal tool access, the Execution Subagent is sufficient to allow the main agent to complete tasks meaningfully. We also see that Terminus-4B is able to perform this role as effectively as Frontier models at a fraction of the cost.

Without the main agent's ability to call Terminal tool, the impact of subagent quality becomes starkly visible. The Vanilla-4B model actually drives the token usage above the Opus baseline (+9.5\%) and exhibits a Sub$\to$Sub rate of 1.51, meaning the main agent re-invokes the subagent over 1.5 times per instance on average, which is a $\sim$70\% increase over Opus. Another thing we see is that the subagent trajectories are too short (only 18k tokens) indicating they may lack detail needed for the main agent to proceed confidently. In contrast, we see that both SFT and Terminus-4B models improve significantly over the Vanilla model and Opus baseline in terms of Frontier LLM token consumption ($\sim$20\%). We also see that the token consumption of the Subagent trajectory is also higher, evident from the SLM token usage per instance for both SFT and Terminus models.

Comparing the SFT and RL models, we see that the distrust metric of Sub$\to$Sub rate improves from SFT to Terminus-4B model, from over 1.0 to 0.89 per instance, matching Opus itself. This shows that the RL stage is crucial in teaching the model to produce responses that satisfy the main agent on the first attempt.

\begin{figure}[t]
   \centering
    \includegraphics[width=0.95\columnwidth]{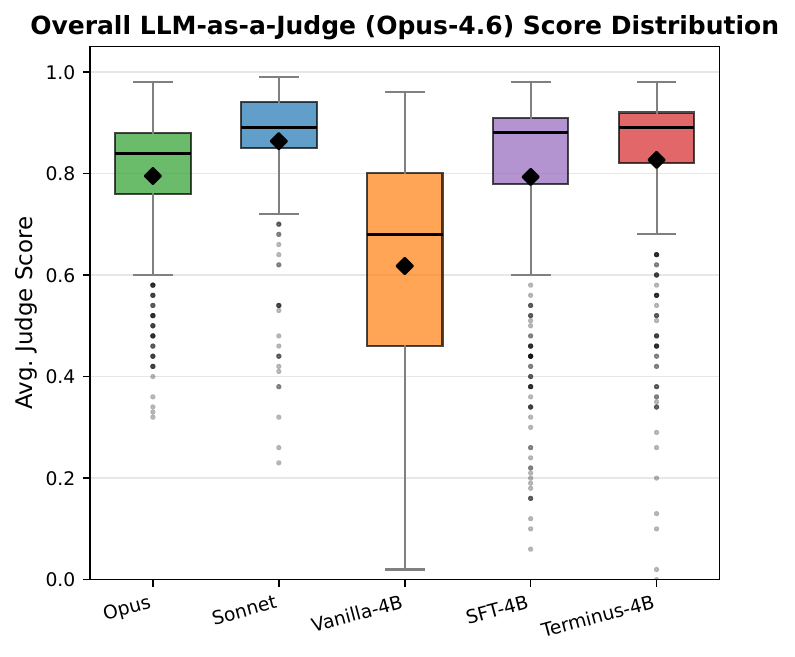}
   \caption{LLM judge scores for subagent response quality from the main agent's perspective in the No Terminal ablation. We show the overall score distributions across the different subagent models.}
   \label{judge_scores}
\end{figure}

\subsubsection{Assessing Response Quality via an LLM-Judge}
To measure the utility of the subagent response from the main agent's perspective, we also employ an LLM-as-a-Judge evaluation. We use Claude Opus-4.6 as the Judge model and prompt it to score the subagent response given the trajectoy leading up to the subagent invocation and N steps following it in the main agent trajectory. We set N to 5 for our evaluation as we believe that should be sufficient turns to observe main agent's ability to use the response. We prompt the Judge LLM to score the subagent response along 5 dimensions (each scored from 0 to 1):
\begin{itemize}
    \item \textbf{Task Completion}: Did the subagent fully accomplish the task it was asked to complete?
    \item \textbf{Factual Accuracy}: Are the claims made by the subagent grounded in fact or does it hallucinate?
    \item \textbf{Informativeness}: Does the response contain sufficient detail for the main agent to proceed without re-running commands?
    \item \textbf{Relevance}: Is the response focused on the task or is it doing unnecessary work outside of it?
    \item \textbf{Actionability}: Can the main agnet figure out what to do next based on the response? Is the output of the subagnet actionable?
\end{itemize}

The overall score is the mean across all five dimensions. We run this evaluation on the No Terminal tool ablation runs. As input the Judge prompt includes the following information: 1) the main agent's system prompt and original problem statement, 2) the trajectory leading up to the subagent call (prior tool calls and results), 3) the subagent query and response and 4) the subsequent 5 turns of the main agent's trajectory after receiving the subagent response. The subseqeunt trajectory is key here because it provides the judge the ability to understand how the main agent responds to the subagent response and whether it's able to use it readily for does it have to re-invoke the subagent with a different query.

Looking at the results, we can see that Terminus-4B achieves scores comparable to frontier LLMs. Interestingly, we see that Sonnet actually performs the best in this evaluation. Terminus-4B results seem to be approaching those of Sonnet and are slightly better than Opus. This corroborates the distrust metric of Sub$\to$Sub being equal between Opus and Terminus-4B in this configuration (as seen in Table~\ref{swebcs_no_terminal}). 

Comparing the scores between the smaller models specifically, we can clearly see that the median score for Terminus-4B is higher than that of SFT-4B. Both SFT and Terminus-4B are better than the Vanilla model, showing the importance of both stages of our post-training.

\subsubsection{A Terminal-Focused Evaluation (using TerminalBench 2.0)}
While SWE-Bench Pro and SWE-Bench C\# evaluate the Execution Subagent in the context of a full repo, TerminalBench 2.0~\cite{terminalbench} provides a complementary evaluation where terminal execution is the \textit{primary} task. The benchmark consists of 89 terminal tasks spanning data processing, system administration, etc. We run the same No Terminal ablation as before, using Opus 4.6 as the main agent, and report metrics over the 75 instances with valid evaluations across all configurations. Table~\ref{terminalbench-eval} shows the results.

An interesting result is that Terminus-4B \textit{improves} the resolve rate slightly to 62.7\%, a gain of +4\% over the No Subagent and Opus baselines. However, since this is over only 75 valid instances, this could be attributed to some fluctuation in agent behavior. In terms of token efficiency, Terminus-4B reduces frontier tokens by $\sim$47\%, about 15\% more compared to Opus. These savings are also notably larger than those observed on SWE-Bench Pro and SWE-Bench C\#, likely because this benchmark is terminal heavy so the heavy-lifting is being done by the subagent.

Looking at the behavioral metric, Sub$\to$Sub, we see the Vanilla-4B model shows the weakest performance with the highest Sub$\to$Sub value among all configurations. This means that main agent is issueing back to back subagent calls, likely having to redo work. Further, we also also see low subagent token usage (24k) indicating insufficiently detailed responses. Both SFT and RL models bring Sub$\to$Sub down significantly (1.03 and 1.07 respectively), even outperforming Opus (1.15). This indicates that the main agent trusts and acts on Terminus-4B's responses without needing to re-invoke the subagent, matching the pattern we observed in the SWE-Bench C\# No Terminal ablation.

\begin{table*}[t]
\centering
\caption{Results on TerminalBench 2.0 with Terminal tool removed, using Claude Opus 4.6 as the main agent model. All terminal work must flow through the Execution Subagent. Metrics are computed over the 75 instances with valid evaluations across all configurations. We show percentage changes relative to the No Subagent baseline. For Frontier tokens, we include main agent tokens plus subagent tokens when the subagent is a frontier model. SLM tokens are consumed only by the 4B subagents.}
\label{terminalbench-eval}
\scriptsize
\begin{tabular}{lclclcclcc}
\toprule
\textbf{Subagent} & \textbf{Resolve \%} & \textbf{Main Agent Tokens} & \textbf{Subagent Tokens} & \textbf{Frontier LLM Tokens} & \textbf{SLM Tokens} & \textbf{Sub Calls} & \textbf{Sub$\to$Sub} & \textbf{Final Ans.\ \%} \\
\midrule
No Subagent  & 58.7 & 471k & - & 471k & - & - & - & - \\
Opus         & 58.7 & 259k (\impact{-44.9\%}{green}) & 60k & 319k (\impact{-32.3\%}{green}) & - & 3.2 & 1.15 & 100.0 \\
Sonnet       & 57.3 & 281k (\impact{-40.4\%}{green}) & 55k & 336k (\impact{-28.7\%}{green}) & - & 2.7 & 1.14 & 100.0 \\
Vanilla-4B   & 60.0 & 374k (\impact{-20.7\%}{green}) & 24k & 374k (\impact{-20.7\%}{green}) & 24k & 4.6 & 1.68 & 97.6 \\
SFT-4B       & 60.0 & 242k (\impact{-48.6\%}{green}) & 58k & 242k (\impact{-48.6\%}{green}) & 58k & 3.1 & 1.03 & 98.6 \\
Terminus-4B       & 62.7 & 251k (\impact{-46.7\%}{green}) & 50k & 251k (\impact{-46.7\%}{green}) & 50k & 3.1 & 1.07 & 97.8 \\
\bottomrule
\end{tabular}
\end{table*}

\section{Limitations}
While our results demonstrate that finetuned small language model can effectively replace Frontier LLMs at the task of terminal execution in coding agents, there remain several limitations:

\textbf{Platform / Shell Coverage.}
Both our training and evaluation are skewed towards Unix / Bash-based terminal tasks. We do not include tasks from other shells like Powershell or Command Prompt for Windows or Zsh for Mac-based usage, despite these being common in real-world development. Extending Terminus-4B to handle these other shells is a natural future direction for this work.

\textbf{Evaluation Scope.}
Our evaluation is almost exclusively conducted on SWE-Bench style benchmarks, which are drawn from GitHub issues and are solved by running the agent in Docker containers already containing the dependencies needed for the project. While this is consistent with prior coding agent evaluation in literature, this may not reflect real world usage of agents, which may be more convoluted and messy. It may feature a myriad of tasks ranging from debugging the application to deployment or infrastructure-level tasks.

\textbf{Base Model Choice.}
We train exclusively on Qwen3-4B model, which represents a reasonable size and family of models for this task. Our results demonstrates that a 4B-parameter model with sufficient training can prove competent for this task, but our study leaves out larger models (8B, 30B, etc.) as well as alternative model families. It remains to be seen whether the same post-training recipe transfers effectively to other families / model sizes.

\section{Conclusion}
In this work, we presented the Execution Subagent and Terminus-4B, a 4B-parameter language model fine-tuned to serve as the execution subagent within coding agents. Through a two-stage post-training pipeline, we showed that even a small language model can match or exceed Frontier LLM performance at a small contained task of terminal execution, while \textbf{reducing the frontier LLM token usage by up to $\sim$30\%}. Our extensive evaluation on benchmarks like SWE-Bench Pro, TerminalBench 2.0 and SWE-Bench C\#, demonstrates that Terminus-4B generalizes across programming languages and choices of main agent models spanning two popular model families (Claude and GPT). The behavioral metrics across our evaluation confirm that the post-training significantly improves on the Vanilla model (Qwen3-4B) and the main agent learns to rely on Terminus-4B responses at the same level as Frontier LLM, rarely needing to redo work. Broadly, our work also demonstrates a practical paradigm of training smaller LMs as subagents and using them to reduce the cost of running coding agents, by diverting token usage to the subagent. The subagent architecture provides a natural way to decompose complex tasks in smaller sub-tasks which can be shared across models of differing capabilities. We believe that our results are directly applicable to other subagent types and provide a scalable path towards coding agents that are more capable, cost-effective and accessible to everyone.

\bibliographystyle{IEEEtran}
\bibliography{iclr2024_conference}

\end{document}